\documentclass[conference]{IEEEtran}
\IEEEoverridecommandlockouts
\usepackage{cite}
\usepackage{amsmath,amssymb,amsfonts}
\usepackage{algorithmic}
\usepackage{graphicx}
\usepackage{textcomp}
\usepackage{xcolor}
\usepackage{hyperref}
\usepackage{booktabs}
\usepackage{pifont}     
\usepackage{makecell}   
\usepackage{threeparttable}
\usepackage{pgfplots}
\usepackage{stfloats}
\usepackage{cuted}
\usepackage{capt-of}
\fnbelowfloat

\pgfplotsset{compat=1.18}
\def\BibTeX{{\rm B\kern-.05em{\sc i\kern-.025em b}\kern-.08em
    T\kern-.1667em\lower.7ex\hbox{E}\kern-.125emX}}

\renewcommand{\figurename}{Figure}
\newcommand{\modelname}{\textsc{Weave}}

\begin{document}

\title{
\modelname{}: Learning Whole-Body Dexterous Loco-Manipulation from Human--Object Interactions
}

\author{
\IEEEauthorblockN{
    Liu Cao\textsuperscript{1,$\dagger$},
    Xingze Wu\textsuperscript{3,$\dagger$},
    Jingzhi Cui\textsuperscript{1},
    Botian Xu\textsuperscript{4},
    Mingzhi Pei\textsuperscript{2},
    Ruoqu Chen\textsuperscript{1,$\dagger$},
    Mengdi Xu\textsuperscript{1}%
    }
    \IEEEauthorblockA{
        \textsuperscript{1}Tsinghua IIIS\qquad
        \textsuperscript{2}Tsinghua College AI\\
        \textsuperscript{3}Dalian University of Technology\qquad
        \textsuperscript{4}The Chinese University of Hong Kong
    }
}

\maketitle

\begin{figure*}[!b]
  \centering
  \includegraphics[width=\textwidth]{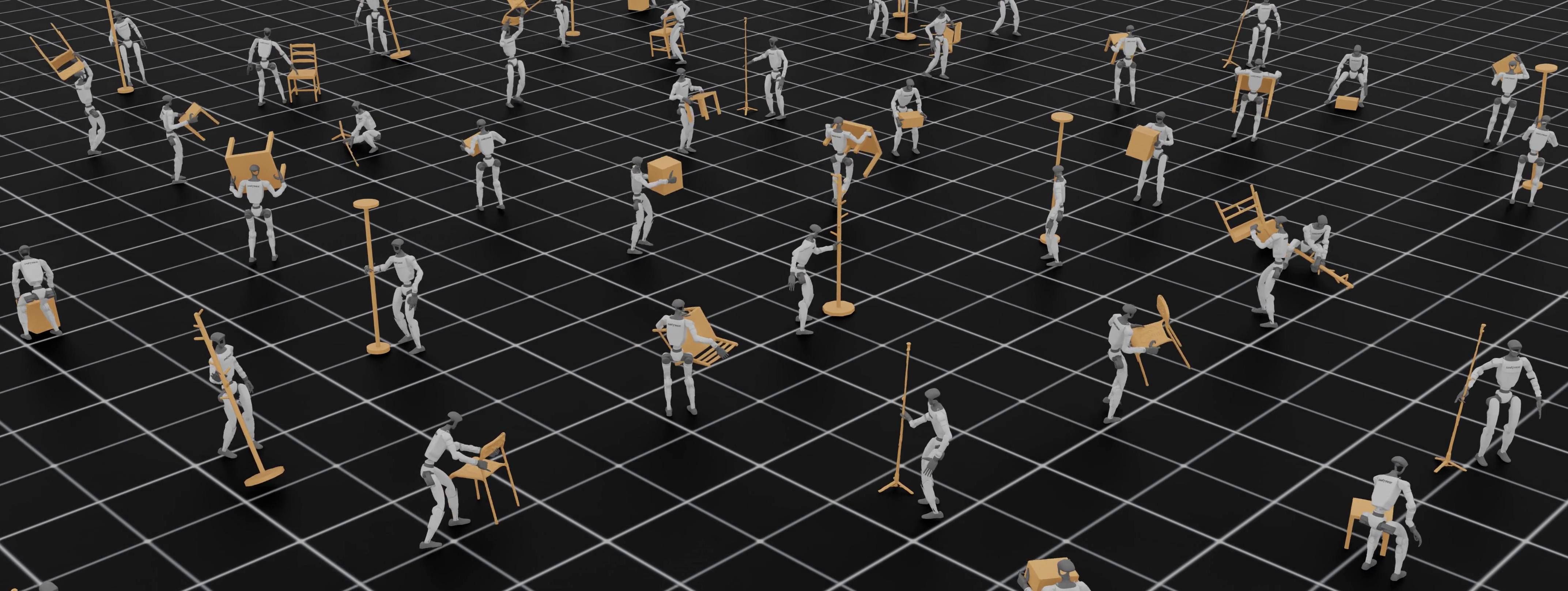}
  \caption{Policy rollout examples with \textbf{\modelname{}.} 
  We learn a unified policy for whole-body dexterous loco-manipulation from captured human--object interaction.
  Parallel simulation rollouts showcase the one policy coordinating
  locomotion and dexterous manipulation across diverse everyday objects.}
  \label{fig:teaser}
  \vspace{4pt}
  {\raggedright\rule{0.4\columnwidth}{0.4pt}\\[2pt]
  \footnotesize $\dagger$Work partially done at CocoMatrix.\par}
\end{figure*}

\begin{abstract}
Learning humanoid--object interaction requires coordinating whole-body balance, locomotion, and dexterous hand contact to control both robot and object motion. 
Human demonstrations provide examples of coordinated interaction, but transferring these behaviors to humanoid robots requires learning how to establish and maintain effective contacts under different embodiments and dynamics. 
We present \modelname{}, a unified framework for learning whole-body dexterous humanoid--object interaction from captured human demonstrations. 
\modelname{} first converts captured human--object interactions into executable robot--object references through contact-aware retargeting and approach-motion completion.
At its core is a contact- and geometry-aware policy that jointly commands $29$ body joints and $12$ actuated finger joints across multiple objects and interaction sequences.
Evaluation across nine objects yields a $92.5\%$ success rate on trained interactions and, without any additional training, $65.0\%$ on sequences never seen during training.
We additionally release $\sim9{,}000$ physically executed rollouts spanning $\sim$23 hours, providing robot--object trajectories with contact annotations for downstream interaction-policy learning and physically consistent HOI motion generation.
Our project website: \href{https://xiaohu-art.github.io/Weave/}{https://xiaohu-art.github.io/Weave/}.
\end{abstract}

\section{Introduction}
\label{sec:intro}

Humanoid robots are expected to operate in environments built for humans, where many everyday tasks require the robot to manipulate objects---approaching, grasping, and transporting them while maintaining balance. Such contact-rich loco-manipulation is a core capability for general-purpose humanoids. Even a task as simple as moving a chair requires the robot to establish a stable multi-finger grasp, generate sufficient contact forces, and continuously adapt its whole-body posture to the resulting load. However, learning such behaviors with a unified policy remains challenging, as balance, locomotion, object motion, and dexterous grasping are tightly coupled: an inaccurate grasp alters the forces acting on the upper body, while poor whole-body coordination can destabilize both the robot and the object.

A primary challenge is \textbf{learning contact-rich robot-object interactions from human demonstrations}. 
Reference motion tracking~\cite{liu2025opt2skill,weng2025hdmi,yang2025omniretarget,zhao2025resmimic,chen2026scenebot} offers a route from demonstrations to physically feasible behavior, but supervises robot and object motion at the body and arm level and leaves finger-level hand--object contact unmodeled; dexterous interaction learning~\cite{luo2024omnigrasp,luo2025emergent,tessler2025maskedmanipulator,xu2025intermimic,xu2026interprior,wu2025human} does model such contact, yet targets simulated characters or fixed-base hands whose kinematics and actuation differ from a humanoid with underactuated hands.
Human demonstrations describe coordinated body and object motion, however, differences in morphology and actuation change how the humanoid must establish hand--object contact, support the object, and maintain balance. 
The resulting control problem is also high-dimensional and heterogeneous: body joints govern balance and load bearing on timescales set by locomotion, while actuated finger joints govern contacts whose outcome turns on millimeter-scale placement, and both must be commanded by one policy.
Even for a single interaction, matching demonstrated poses does not ensure that the robot can sustain the force and torque required to move the object. 
Learning therefore requires coupling robot--object motion tracking with the acquisition of contact-aware, closed-loop control under the robot's dynamics.

A secondary challenge is \textbf{learning diverse object interactions within a unified policy}.
Across objects, differences in geometry affect the feasible contact locations and grasp configurations. 
Across interaction sequences, variations in approach direction, body posture, and object motion require different control behaviors, even for the same object.
Prior dexterous interaction methods largely train one policy per object or per clip~\cite{xu2025intermimic}, which sidesteps this variation.
The challenge is to learn a shared control strategy that captures common coordination patterns across interactions while adapting whole-body motion and finger-level contact to the specific object and reference sequence.

In this work, we investigate how to equip humanoids with whole-body dexterous interaction capability from captured human demonstrations. 
Our key idea is to make object geometry and hand--object contact explicit throughout the pipeline: references are constructed to preserve task-relevant contacts across the morphology gap, and a unified policy, conditioned on object geometry and trained jointly across objects and interaction clips, tracks robot and object motion while explicitly learning whole-body coordination and finger-level contact.

We present \modelname{}, an end-to-end framework for learning whole-body dexterous loco-manipulation from human--object interaction. 
Given captured human--object interactions, which records full-body human motion while manipulating everyday objects such as chairs, boxes, and tables, \modelname{} completes the missing approach and transition motions with whole-body motion generation~\cite{rempe2026kimodo}, retargets the interaction to the robot with contact-aware retargeting, and trains the contact- and geometry-aware policy in simulation.
As a result, the robot walks up to an object, grasps it with multi-fingered hands, lifts it, and carries it to a target configuration.
We evaluate the learned policy and publicly release a curated dataset of simulation rollouts to support research on whole-body dexterous interaction and learning from robot demonstrations.

In summary, our contributions are as follows: \begin{itemize}
    \item \textbf{Contact- and geometry-aware whole-body dexterous policy.} A unified policy, trained jointly across multiple objects and interaction clips, that tracks robot and object motion while explicitly learning hand--object contact, reproducing whole-body coordination, object transport, and finger-level grasping within one policy. 
    \item \textbf{Interaction-preserving trajectory construction.} A pipeline that converts captured human--object interactions into executable robot--object trajectories through approach-motion completion and contact-aware retargeting, preserving object motion and hand--object contacts across the human--robot morphology gap.
    \item \textbf{Systematic evaluation and physically plausible dataset.}
    We conduct extensive experiments to assess interaction execution, quantify the benefits of joint multi-object learning, and examine how policy architecture and optimization affect learning efficiency, together with the resulting rollout dataset released for downstream policy learning and humanoid-object interaction modeling.
\end{itemize}

\section{Related Works}

\subsection{Interaction-Preserving Retargeting and Motion Generation} Motion retargeting constructs reference motions from human demonstrations, conventionally by optimizing joint configurations to match Cartesian keypoints~\cite{luo2023perpetual, Zakka_Mink_Python_inverse_2026, he2024omnih2o}. For human--object interaction, however, pose similarity alone is insufficient: embodiment differences can alter hand--object contacts, break the relative interaction geometry, or produce configurations that cannot exert the intended effect on the object. Interaction-preserving retargeting therefore incorporates contact constraints: OmniRetarget~\cite{yang2025omniretarget} preserves whole-body interaction structure for humanoids, TopoRetarget~\cite{wu2026toporetarget} and REGRIND~\cite{feng2026minimalist} encode local hand--object topology for dexterous hands, and other works~\cite{wu2026reforce, shao2026synmandex, zhu2026learning, chen1dex} further enforce force consistency of the retargeted contacts. In parallel, human motion generation has progressed from parametric motion synthesis~\cite{wang2026motionbricks} to more scalable and controllable models: Kimodo~\cite{rempe2026kimodo} handles heterogeneous spatial and temporal constraints for both human and humanoid embodiments, and ARDY~\cite{zhao2026ardy} extends to autoregressive diffusion for streaming generation. 
\modelname{} combines contact-aware retargeting with Kimodo-based approach-motion completion to construct robot--object reference trajectories. 
These kinematic references then supervise reinforcement learning, which converts them into physically executable interactions in simulation.

\subsection{Humanoid Loco-Manipulation} 
Humanoid loco-manipulation requires coordinating locomotion, balance, and object interaction over extended task horizons. Modular approaches organize navigation, locomotion, reaching, and manipulation into separate components~\cite{chen2025hand,arnaud2025locate,liu2025compass,he2025viral}. Such decomposition provides structured interfaces, while coordinating balance and hand--object contact across these interfaces remains an important consideration. 
Reference-tracking approaches instead learn a physics-based policy that tracks diverse humanoid motions and object interactions~\cite{liu2025opt2skill,weng2025hdmi,yang2025omniretarget,Taouil2026MotionDiscoMD,zhao2025resmimic,chen2026scenebot}. 
Physics-based grasping and human--object interaction
methods~\cite{luo2024omnigrasp,tessler2025maskedmanipulator,xu2025intermimic,xu2026interprior,wu2025human}
highlight the importance of coupling motion tracking with
hand--object contact, motivating unified control across
objects and interaction sequences.
A parallel line of work introduces egocentric depth or RGB observations for visually guided loco-manipulation~\cite{li2026vaic,wang2025physhsi,he2026ultra,byrd2026adaptmanip,yin2025visualmimic}. Data-generation approaches further diversify humanoid interactions: GRAIL~\cite{xie2026grailgeneratinghumanoidlocomanipulation} synthesizes 4D humanoid--object interactions with a video generation model~\cite{team2026kling}, HumanoidMimicGen~\cite{lin2026humanoidmimicgen} augments a few teleoperated demonstrations through whole-body planning, and VLK~\cite{wang2026vlk} synthesizes paired vision--language--kinematics data in reconstructed scenes. 
Relative to this literature, \modelname{} supervises finger-level contact rather than body- and arm-level motion alone, learns a single policy across objects and interaction clips rather than per-object controllers, and releases the executed rollouts as reusable interaction data.

\subsection{Humanoid Vision--Language--Action Models}
\label{sec:h-vla}
VLA models connect semantic task descriptions and visual observations with executable robot actions. General-purpose systems~\cite{kim2024openvla,black2024pi_0,intelligence2025pi_,bjorck2025gr00t} leverage large-scale robot datasets and pretrained VLM representations to generalize across scenes and tasks. Most of these models were initially developed for fixed-base manipulators, whose action spaces and stability requirements differ substantially from those of humanoid robots.

Extending VLA models to humanoids requires reasoning over high-dimensional whole-body motion while maintaining balance and physical contact. 
Recent humanoid-specific systems~\cite{jiang2025wholebodyvla,bai2026hex,wei2026psi_0,li2026omega}, explore unified or hierarchical architectures that connect semantic reasoning with whole-body motor control. Nevertheless, acquiring large-scale data that jointly contain locomotion, precise object interaction, and articulated finger motion remains substantially more difficult than collecting arm-level manipulation trajectories.

Therefore, \modelname{} is complementary rather than competing: instead of collecting interactions through teleoperation or generation, we acquire the skills through physics-based reinforcement learning on human-object motion, producing grounded, finger-level behavior.
Such physically grounded skills can serve as reusable low-level capabilities for future humanoid VLA systems, while VLA models can provide the semantic task specifications and high-level motion commands needed to select and compose them.

\begin{figure*}[t]
  \centering
  \includegraphics[width=\textwidth]{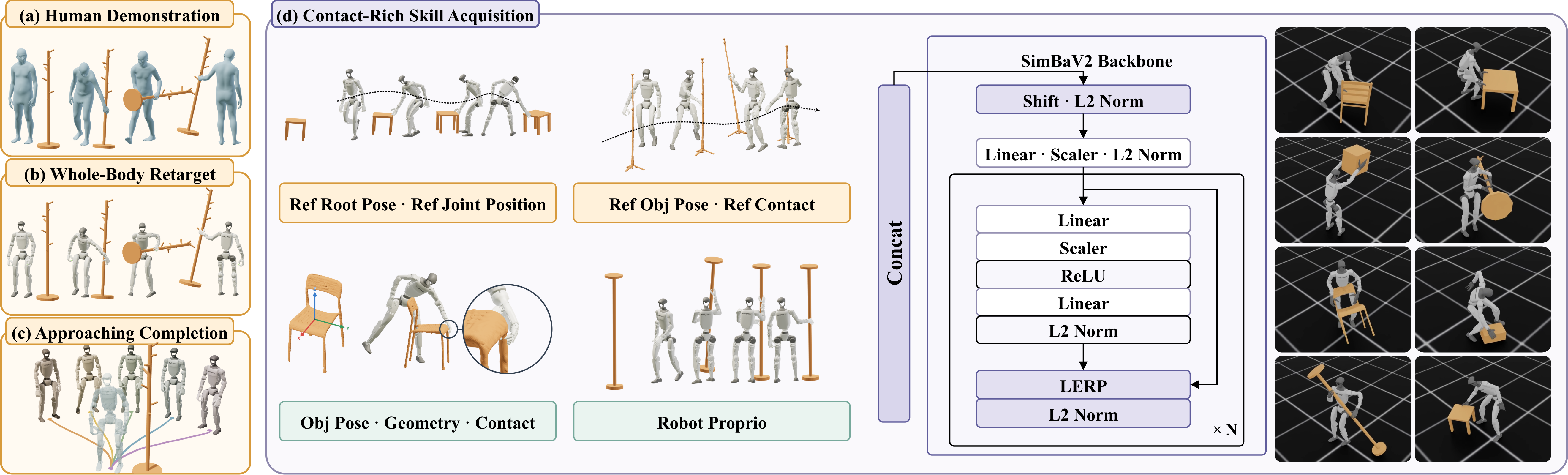}
  \caption{\textbf{Overview of \modelname{}.}
  \textbf{(a)} Captured human--object interactions provide paired SMPL-X body motion and object trajectories.
  \textbf{(b)} Whole-body inverse kinematics retargets each interaction while preserving the hand--object contacts and the object motion.
  \textbf{(c)} Kimodo~\cite{rempe2026kimodo} synthesizes locomotion prefixes that approach the initial interaction pose from different azimuths, producing several complete references from one captured sequence.
  \textbf{(d)} A unified reference-conditioned policy is trained in simulation across all objects and interaction clips.}
  \label{fig:method}
\end{figure*}

\section{Problem Formulation}
\label{sec:problem}

We study whole-body dexterous loco-manipulation, in which a humanoid equipped with dexterous hands must approach an object, establish a stable grasp, and transport the object toward a desired configuration while maintaining whole-body balance. We formulate the closed-loop control problem as a partially observable Markov decision process (POMDP)
$\mathcal{M}=(\mathcal{S},\mathcal{A},\mathcal{O},\mathcal{T},r,\gamma)$,
where $\mathcal{S}$ and $\mathcal{O}$ denote the state and observation spaces, $\mathcal{A}$ is the action space, $\mathcal{T}$ is the state-transition function, $r$ is the reward function, and $\gamma$ is the discount factor. At each time step $t$, the policy outputs an action
$a_t \sim \pi_\theta(\cdot\mid o_t)$,
where $a_t$ specifies desired joint positions for low-level PD controller.

Policy learning is bootstrapped from captured human--object interaction data~\cite{li2023objectmotionguidedhuman}. After retargeting and motion completion, each sequence provides a robot--object reference trajectory
\begin{equation}
\tau^{\mathrm{ref}}
=
\left\{
\hat{s}^{\mathrm{ref}}_t
\right\}_{t=0}^{T-1},
\qquad
\hat{s}^{\mathrm{ref}}_t
=
\left(
\hat{x}^{\mathrm{robot}}_t,
\hat{x}^{\mathrm{obj}}_t
\right),
\label{eq:reference_trajectory}
\end{equation}
where $\hat{x}^{\mathrm{robot}}_t$ specifies the reference humanoid configuration and $\hat{x}^{\mathrm{obj}}_t$ specifies the corresponding object pose. The reference trajectory is used to initialize state in simulation and construct the tracking objectives during policy training.

\section{Method}
\label{sec:method}

\subsection{Overview}
\modelname{} transforms captured human--object interactions into whole-body dexterous loco-manipulation skills through two main stages. First, we construct reference trajectories using whole-body retargeting, contact-aware hand refinement, and approach-motion completion. 
Then we train a unified contact- and geometry-aware policy through reference-tracking, jointly coordinating the humanoid body, articulated fingers, and object motion across multiple objects and interaction clips.

\subsection{Interaction-Preserving Reference Construction}
\label{sec:reference_construction}

Before policy learning, we convert captured human--object interaction sequences into reference motions. 
Starting from the paired SMPL-X motion and object trajectory, our pipeline first retargets the interaction segment to the humanoid through whole-body inverse kinematics and contact-aware hand refinement. 
We then use Kimodo~\cite{rempe2026kimodo} to synthesize a locomotion prefix that approaches the initial interaction pose. This process produces a synchronized robot--object trajectory spanning approach, grasping, and object transport, together with contact annotations used for policy training.

\textbf{Whole-body retargeting.}
We first obtain an initial robot trajectory
$\{\tilde{x}^{\mathrm{robot}}_t\}_{t=0}^{T-1}$
using a GMR-style whole-body inverse-kinematics objective~\cite{araujo2025retargeting}, which aligns selected robot links with their SMPL-X counterparts. To account for morphological differences, we constrain the pelvis only in the horizontal plane and optimize its height based on ground contact in place of scale calibration, while joint-limit, velocity, and acceleration penalties promote feasible and temporally smooth motion.

\begin{table*}[!ht]
\centering
\small
\setlength{\tabcolsep}{4pt}
\renewcommand{\arraystretch}{1.2}
\caption{Reward terms for reference tracking. Hatted quantities denote
references. Tracking terms contribute
$w_i\exp(-e_{t,i}/\sigma_i^2)$.}
\label{tab:stage1_rewards}
\begin{tabular*}{\textwidth}{@{\extracolsep{\fill}}lllcc@{}}
\toprule
Reward term & Computation & Definition & Weight $w$ & Scale $\sigma$ \\
\midrule
\multicolumn{5}{@{}l}{\textit{Robot and object tracking}} \\

Pelvis position
&
$\lVert p^r_t-\hat p^r_t\rVert_2^2$
&
$p^r_t,\hat p^r_t$: current/reference pelvis position
&
$1.0$ & $0.3$
\\

Pelvis orientation
&
$\angle(R^r_t,\hat R^r_t)^2$
&
$R^r_t,\hat R^r_t$: pelvis orientation; $\angle$: geodesic error
&
$1.0$ & $0.4$
\\

Body positions
&
$\displaystyle
\frac{1}{|\mathcal B|}
\sum_{b\in\mathcal B}
\lVert p_{t,b}-\hat p_{t,b}\rVert_2^2$
&
$\mathcal B$: tracked robot links
&
$1.0$ & $0.3$
\\

Body orientations
&
$\displaystyle
\frac{1}{|\mathcal B|}
\sum_{b\in\mathcal B}
\angle(R_{t,b},\hat R_{t,b})^2$
&
$R_{t,b},\hat R_{t,b}$: current/reference link orientation
&
$1.0$ & $0.4$
\\

Object position
&
$\lVert p^o_t-\hat p^o_t\rVert_2^2$
&
$p^o_t,\hat p^o_t$: current/reference object position
&
$2.0$ & $0.2$
\\

Object orientation
&
$\angle(R^o_t,\hat R^o_t)^2$
&
$R^o_t,\hat R^o_t$: current/reference object orientation
&
$2.0$ & $0.3$
\\

\midrule
\multicolumn{5}{@{}l}{\textit{Grasp and contact}} \\

Hand opposition
&
$\displaystyle
\frac{
\sum_h \eta_{t,h}
\left(
\frac{1}{|\mathcal K_h|}
\sum_{k\in\mathcal K_h}
\frac{1-u_{t,h,0}^{\top}u_{t,h,k}}{2}
\right)
}{
\sum_h \eta_{t,h}+\epsilon
}$
&
\begin{tabular}[c]{@{}l@{}}
$u_{t,h,k}$: surface-to-fingertip unit vector;\\
$k=0$: thumb; \\
$\eta_{t,h}$: expected-contact gate
\end{tabular}
&
$2.0$ & --
\\

Contact matching
&
$\displaystyle
\frac{
\sum_{h\in\mathcal H}
m_{t,h}(1-|y_{t,h}-\tilde c_{t,h}|)
}{
\sum_{h\in\mathcal H}m_{t,h}+\epsilon
}$
&
\begin{tabular}[c]{@{}l@{}}
$m_{t,h}=\mathbf{1}[\hat c_{t,h}\neq0]$;\\
$y_{t,h}=\mathbf{1}[\hat c_{t,h}>0]$;\\
$\tilde c_{t,h}=\min(\lVert f_{t,h}\rVert/\bar f,1)$
\end{tabular}
&
$2.0$ & --
\\

\midrule
\multicolumn{5}{@{}l}{\textit{Regularization}} \\

Foot sliding
&
$\displaystyle
\sum_{f\in\mathcal B_{\mathrm{foot}}}
\chi^{\mathrm{ground}}_{t,f}
\lVert v^{xy}_{t,f}\rVert_2$
&
$\chi^{\mathrm{ground}}_{t,f}$: foot--ground contact indicator
&
$-1.0$ & --
\\

Action rate
&
$\lVert a_t-a_{t-1}\rVert_2^2$
&
$a_t,a_{t-1}$: current/previous joint command
&
$-0.1$ & --
\\

Joint limits
&
$\displaystyle
\sum_{j\in\mathcal J_{\mathrm{body}}}
\left(
[q_{t,j}-\overline q_j]_+
+
[\underline q_j-q_{t,j}]_+
\right)$
&
\begin{tabular}[c]{@{}l@{}}
$\mathcal J_{\mathrm{body}}$: non-finger joints;\\
$[\underline q_j,\overline q_j]$: soft joint limits
\end{tabular}
&
$-10.0$ & --
\\

\bottomrule
\end{tabular*}
\end{table*}

\textbf{Contact-aware hand refinement.}
Whole-body IK captures the arm and wrist motion but does not ensure a stable fingertip grasp.
We therefore jointly refine the arm and finger configurations of both limbs,
$\xi_t=(q_t^{\mathrm{arm}},q_t^{\mathrm{hand}})$,
initialized from $\{\tilde{x}^{\mathrm{robot}}_t\}_{t=0}^{T-1}$ and held fixed elsewhere: the pelvis pose and the leg configuration remain at their IK solution, so the refinement does not alter the retargeted whole-body motion.
For fingertip $k$ at frame $t$, let
$p_{t,k}(\xi_t)$ be its forward-kinematics position,
$\bar p_{t,k}$ the object-surface point nearest to the corresponding human fingertip,
$n_{t,k}$ its outward normal, and
$c_{t,k}\in\{0,1\}$ the contact label.
Following~\cite{shao2026synmandex}, we minimize
\begin{equation}
\begin{aligned}
\mathcal{L}_{\mathrm{hand}}
={}&
\underbrace{
\lambda_a \sum_{t,k} c_{t,k}
\lVert p_{t,k}-\bar p_{t,k}\rVert
}_{\text{contact attraction}}
-
\underbrace{
\lambda_q \sum_t
\hat Q_{\mathrm{FC}}(q_t^{h})
}_{\text{force-closure objective}}
\\
&+
\underbrace{
\lambda_p \sum_{t,k}
\left[-(p_{t,k}-\bar p_{t,k})^{\top}n_{t,k}\right]_{+}
}_{\text{penetration penalty}}
+
\ \mathcal{L}_{\mathrm{reg}} .
\end{aligned}
\label{eq:hand_obj}
\end{equation}
where $[z]_{+}=\max(z,0).$
The attraction term alone only places fingertips on the object surface, which admits configurations that touch the object without being able to hold it. We therefore additionally require the contacts to resist arbitrary disturbance wrenches, and maximize a differentiable approximation $\hat Q_{\mathrm{FC}}$ of the force-closure quality~\cite{ferrari1992planning},
\begin{equation}
Q_{\mathrm{FC}}(\xi_t)
=
\min_{\lVert w\rVert_2=1}
\max_{\substack{
f\in\mathcal{F}(\xi_t)\\
\lVert f\rVert_1\leq 1}}
w^{\top}G(\xi_t)f.
\label{eq:qfc}
\end{equation}
Here, $G$ maps feasible contact forces $f$ to object-centric wrenches, and $\mathcal{F}$ is the linearized friction cone with soft-finger torsion. Intuitively, $Q_{\mathrm{FC}}$ measures the weakest disturbance wrench that the grasp can resist; $Q_{\mathrm{FC}}>0$ indicates force closure. We approximate it by a fixed set of wrench directions.

\textbf{Approach motion completion.}
\label{sec:approach}
The retargeted interaction begins near the object and therefore lacks the locomotion required to approach it. We use Kimodo~\cite{rempe2026kimodo} to generate an approach prefix
$\tau^{\mathrm{pre}}$
for the refined interaction trajectory
$\tau^{\mathrm{int}}$.
The initial frames of $\tau^{\mathrm{int}}$ provide terminal whole-body constraints, while the robot follows a path ending at the initial pose and heading. We concatenate the two segments:
\begin{equation}
\tau^{\mathrm{ref}}
=
\tau^{\mathrm{pre}}
\oplus
\tau^{\mathrm{int}},
\qquad
T=T_{\mathrm{pre}}+T_{\mathrm{int}}.
\label{eq:completed_reference}
\end{equation}
Sampling different approach azimuths produces multiple complete references from the same interaction sequence.
We sample $K=3$ for training set and $K=5$ for test set, which is the main source of the trajectory counts reported in \ref{sec:exp_setup}.

\textbf{Reference annotations.}
After concatenation, the completed reference is re-indexed as
$\tau^{\mathrm{ref}}
=\{\hat{s}^{\mathrm{ref}}_t\}_{t=0}^{T-1}$,
where
\begin{equation}
\hat{s}^{\mathrm{ref}}_t
=
\left(
\hat{x}^{\mathrm{robot}}_t,
\hat{x}^{\mathrm{obj}}_t
\right)
=
\left(
\hat{p}^{r}_t,
\hat{R}^{r}_t,
\hat{q}_t,
\hat{p}^{o}_t,
\hat{R}^{o}_t
\right).
\label{eq:reference_state}
\end{equation}
Here, $\hat{p}^{r}_t$, $\hat{R}^{r}_t$, and $\hat{q}_t$ denote the humanoid root position, root orientation, and joint configuration, while $\hat{p}^{o}_t$ and $\hat{R}^{o}_t$ denote the object pose. For each robot link $\ell$, we additionally assign
\begin{equation}
\hat{c}_{t,\ell}
=
\mathbf{1}
\left[d^{o}_{t,\ell}<\delta_{\mathrm{contact}}\right]
-
\mathbf{1}
\left[d^{o}_{t,\ell}>\delta_{\mathrm{far}}\right],
\label{eq:contact_annotation}
\end{equation}
where $d^{o}_{t,\ell}$ is the distance from the link position to the object surface. Thus,
$\hat{c}_{t,\ell}\in\{-1,0,1\}$ denotes separated, neutral, and contact states, respectively.
The completed trajectories provide initialization states and reference motion for policy learning, while the contact annotations specify the desired interaction pattern. Together, they guide a policy that learns to realize the reference interactions under the simulator dynamics.

\subsection{Contact-Rich Skill Acquisition via Reference Tracking}
\label{sec:stage1}

We train a unified reference-conditioned policy in simulation, jointly across multiple objects and interaction clips. The policy uses robot proprioception, object state and geometry features, and reference motion to coordinate whole-body control with finger-level interaction:
\begin{equation}
a_t
\sim
\pi_{\mathrm{track}}
\left(
\cdot
\mid
o^{\mathrm{prop}}_t,
o^{\mathrm{obj}}_t,
\hat{x}^{\mathrm{robot}}_{t:t+H},
\hat{x}^{\mathrm{obj}}_{t:t+H},
\right),
\label{eq:track_policy}
\end{equation}
where $o^{\mathrm{prop}}_t$ is the robot proprioception,
$o^{\mathrm{obj}}_t$ is the simulated object observation, and
$\{\hat{x}^{\mathrm{robot}}_{t:t+H},
\hat{x}^{\mathrm{obj}}_{t:t+H}\}$ 
is a short-horizon command extracted from
$\tau^{\mathrm{ref}}$.
We optimize the policy with PPO~\cite{schulman2017proximal}, using reference state initialization and reference-based early termination following~\cite{peng2018deepmimic}.

\textbf{Observation and action.}
All relative poses and geometric features are expressed in robot local frame. The proprioceptive observation contains the base angular velocity, projected gravity, joint positions and velocities, and the previous action. The object observation contains its ground-truth relative pose, fingertip-to-surface vectors, binary object-contact flags, and BPS-SDF descriptor encoding object geometry. 
The reference command
$\{\hat{x}^{\mathrm{robot}}_{t:t+H},
\hat{x}^{\mathrm{obj}}_{t:t+H}\}$ 
contains the joint configuration, pelvis pose, object pose, and contact labels of a short horizon.

We use an asymmetric actor--critic: the actor receives
$(o^{\mathrm{prop}}_t,o^{\mathrm{obj}}_t,
\hat{x}^{\mathrm{robot}}_{t:t+H},
\hat{x}^{\mathrm{obj}}_{t:t+H}
)$,
whereas the critic replaces $o^{\mathrm{prop}}_t$ with a privileged robot observation that additionally contains the base linear velocity and current tracked-body poses. The action $a_t$ specifies joint-position targets tracked by PD controllers at $50$\,Hz. For the underactuated Inspire hands, the policy commands only the proximal finger joints, while the intermediate and distal joints follow fixed mimic couplings.

\textbf{Tracking objective.}
The reward combines robot and object tracking, geometry-aware grasping, contact matching, and motion regularization:
\begin{equation}
\begin{aligned}
r_t
={}&
\sum_{i\in\mathcal{I}}
w_i\exp\left(-e_{t,i}/\sigma_i^2\right)
+
w_{\mathrm{opp}}r^{\mathrm{opp}}_t
+
w_{\mathrm{c}}r^{\mathrm{contact}}_t \\
&+
w_{\mathrm{slide}}r^{\mathrm{slide}}_t
+
w_{\mathrm{act}}r^{\mathrm{act}}_t
+
w_{\mathrm{lim}}r^{\mathrm{lim}}_t ,
\end{aligned}
\label{eq:track_reward}
\end{equation}
where $\mathcal{I}$ includes pelvis pose, tracked-body pose over the body set $\mathcal{B}$, and object pose. The hand-opposition reward $r^{\mathrm{opp}}_t$ encourages the thumb and opposing fingers to lie on different sides of the object surface whenever the reference specifies a grasp following ~\cite{xu2026interprior}.
For hand link $h\in\mathcal{H}$, let
$\hat c_{t,h}$ be the reference contact label and
$\tilde c_{t,h}=\min(\lVert f_{t,h}\rVert/\bar f,1)$
the normalized contact-force magnitude. Contact matching is computed only for non-neutral labels:
\begin{equation}
r^{\mathrm{contact}}_t
=
\frac{
\sum_{h\in\mathcal{H}}
m_{t,h}
\left(
1-\left|y_{t,h}-\tilde c_{t,h}\right|
\right)
}{
\sum_{h\in\mathcal{H}}m_{t,h}+\epsilon
},
\label{eq:contact_reward}
\end{equation}
where
$m_{t,h}=\mathbf{1}[\hat c_{t,h}\neq0]$
and
$y_{t,h}=\mathbf{1}[\hat c_{t,h}>0]$.
The remaining terms penalize foot sliding, action variation, and non-finger joint-limit violations as showed in Table~\ref{tab:stage1_rewards}.

\begin{figure}[h]
  \centering
  \includegraphics[width=\columnwidth]
  {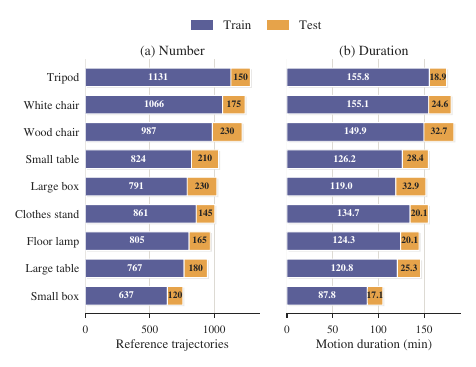}
  \caption{\textbf{Reference-motion dataset distribution.}
  Per-object composition of the training and test splits in terms of
  (a) the number of reference trajectories and (b) total motion
  duration.}
  \label{fig:data_stats}
\end{figure}

\textbf{Randomization and termination.}
Across parallel environments, we randomize robot and object friction and restitution, torso center of mass, and finger actuator properties. Episodes terminate when the pelvis-position error exceeds $0.25$\,m, the object-position error exceeds $0.30$\,m, the projected-gravity error of the pelvis or object exceeds $0.8$ or $0.3$, respectively, or the vertical tracking error of an ankle or wrist exceeds $0.25$\,m. We also terminate after all expected hand--object contacts are absent for ten consecutive control steps. These conditions keep the on-policy state distribution close to the reference~\cite{liao2025beyondmimic,xu2025intermimic}.

\textbf{Policy optimization.}
Each observation group is encoded and projected to latent representation. The concatenated features are processed by SimBaV2 networks~\cite{lee2025simba}. Two-dimensional weight matrices are optimized with Muon~\cite{liu2025muon}, while biases and other parameters use AdamW. We train a unified policy jointly over all objects and reference clips under a $200$\,Hz simulation frequency and a $50$\,Hz control frequency. 

\begin{figure*}[t]
  \centering
  \includegraphics[width=\textwidth]{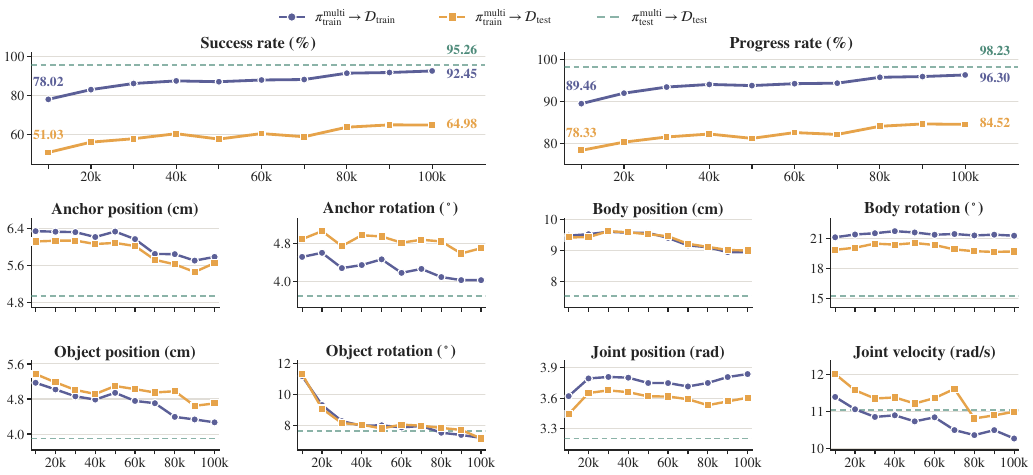}
  \caption{\textbf{Evaluation against training iterations.}
  Success rate (top) and tracking errors (bottom) for
  $\pi^{\mathrm{multi}}_{\mathrm{train}}$ evaluated on
  $\mathcal D_{\mathrm{train}}$ and on $\mathcal D_{\mathrm{test}}$.
  The dashed line is $\pi^{\mathrm{multi}}_{\mathrm{test}}$ evaluated on
  $\mathcal D_{\mathrm{test}}$, which serves as an oracle for measuring generalization.}
  \label{fig:generalization_curves}
\end{figure*}

\section{Experiments}
We evaluate \modelname{} to answer the following questions: 

\textbf{Q1:} How effectively does the learned policy execute whole-body dexterous loco-manipulation across objects and interaction sequences?

\textbf{Q2:} How does joint multi-object training compare with single-object specialist in interaction completion and tracking accuracy?

\textbf{Q3:} How do the policy architecture and optimizer affect learning efficiency?

\subsection{Experimental Setup}
\label{sec:exp_setup}

\textbf{Embodiment and dataset.}
We instantiate \modelname{} on Unitree G1 humanoid with 29 actuated body DoFs and two Inspire dexterous hands with 12 actuated finger DoFs.
We construct two disjoint reference splits,
$\mathcal D_{\mathrm{train}}$ and
$\mathcal D_{\mathrm{test}}$, containing 7,869 (19.56h) and 1,605 (3.67h) reference trajectories respectively as showed in Figure~\ref{fig:data_stats}. 

\textbf{Evaluation metrics.}
For $N$ evaluation clips, let $T_i$ denote the number of executed
frames in clip $i$. We define the clip-balanced average
$
\left\langle z_{i,t}\right\rangle
=
\frac{1}{N}\sum_{i=1}^{N}
\frac{1}{T_i}\sum_{t=0}^{T_i-1} z_{i,t}.
\label{eq:eval_average}
$
The tracking errors are then defined as
\begin{equation}
\begin{gathered}
E_{\mathrm{anchor}}^{p}
=
\left\langle
\lVert p_{i,t}^{a}-\hat p_{i,t}^{a}\rVert_2
\right\rangle,
\quad
E_{\mathrm{anchor}}^{R}
=
\left\langle
\angle(R_{i,t}^{a},\hat R_{i,t}^{a})
\right\rangle,
\\[1mm]
E_{\mathrm{body}}^{p}
=
\left\langle
\frac{1}{|\mathcal B|}
\sum_{b\in\mathcal B}
\lVert p_{i,t,b}-\hat p_{i,t,b}\rVert_2
\right\rangle,
\\[1mm]
E_{\mathrm{body}}^{R}
=
\left\langle
\frac{1}{|\mathcal B|}
\sum_{b\in\mathcal B}
\angle(R_{i,t,b},\hat R_{i,t,b})
\right\rangle.
\end{gathered}
\label{eq:eval_body_errors}
\end{equation}
For joint and object tracking, we report
\begin{equation}
\begin{aligned}
E_{\mathrm{joint}}^{q}
=
\left\langle
\lVert q_{i,t}-\hat q_{i,t}\rVert_2
\right\rangle, & \quad
E_{\mathrm{joint}}^{\dot q}
=
\left\langle
\lVert \dot q_{i,t}-\hat{\dot q}_{i,t}\rVert_2
\right\rangle,\\
E_{\mathrm{obj}}^{p}
=
\left\langle
\lVert p_{i,t}^{o}-\hat p_{i,t}^{o}\rVert_2
\right\rangle, & \quad
E_{\mathrm{obj}}^{R}
=
\left\langle
\angle(R_{i,t}^{o},\hat R_{i,t}^{o})
\right\rangle.
\end{aligned}
\label{eq:eval_joint_object_errors}
\end{equation}
Here, $a$ denotes the pelvis anchor, $\mathcal B$ is the set of tracked
robot bodies, and $\angle(\cdot,\cdot)$ denotes the geodesic rotation
error.

\subsection{Whole-Body Dexterous Loco-Manipulation}
\label{sec:exp_generalization}

\textbf{Protocol.}
We train a multi-object tracking policy
$\pi^{\mathrm{multi}}_{\mathrm{train}}$ on
$\mathcal D_{\mathrm{train}}$ and evaluate it on both
$\mathcal D_{\mathrm{train}}$ and
$\mathcal D_{\mathrm{test}}$
to measure generalization to unseen interaction.
We additionally train $\pi^{\mathrm{multi}}_{\mathrm{test}}$ directly on
$\mathcal D_{\mathrm{test}}$ to serve as a comparison.
The three evaluations are complementary. The first measures how well a single policy fits the references it is trained on, the second measures transfer to unseen interaction sequences of the same objects, and the third serves as an oracle.

\textbf{Results.}
Figure~\ref{fig:generalization_curves} reports both evaluations against training iterations.
Training on $\mathcal D_{\mathrm{train}}$ progresses steadily. The success rate rises from $78.0\%$ at 10k iterations to $92.5\%$ at 100k and the progress rate from $89.5\%$ to $96.3\%$, while the tracking errors decrease over the same span, with object rotation falling from $11.17^\circ$ to $7.18^\circ$ and object position from $5.17$\,cm to $4.27$\,cm. A single policy thus absorbs 7,869 reference trajectories spanning nine objects, and performance is still improving at the end of training.
The policy transfers to unseen interactions, reaching $65.0\%$ SR and $84.5\%$ PR on $\mathcal D_{\mathrm{test}}$ with tracking errors close to those on $\mathcal D_{\mathrm{train}}$, but the gap in completion is substantial. A likely cause is that the two splits are not identically distributed: approach-motion completion generates $K=3$ variations per captured interaction for $\mathcal D_{\mathrm{train}}$ and $K=5$ for $\mathcal D_{\mathrm{test}}$ as introduced in \ref{sec:approach}, so the test split spans a wider range of approach directions and generated locomotion prefixes than the policy encounters during training, leading to a mismatch of data distribution.

\subsection{Joint Multi-Object Skill Learning}
\label{sec:exp_multi_object}

\begin{figure}[h]
  \centering
  \includegraphics[width=\columnwidth]
  {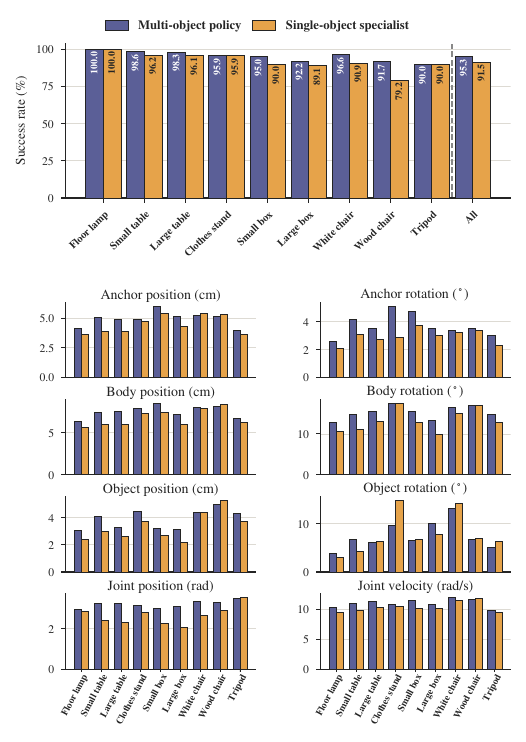}
  \caption{\textbf{Joint multi-object training versus object-specific
  specialists.}
  Top: per-object and pooled success rates on
  $\mathcal D_{\mathrm{test}}$, where the pooled result is weighted by
  the number of evaluation trajectories.
  Bottom: tracking errors computed over rollouts.}
  \label{fig:multi_object_comparison}
\end{figure}

\textbf{Protocol.}
We compare a single multi-object policy $\pi_{\mathrm{test}}^{\mathrm{multi}}$, trained on all objects in $\mathcal D_{\mathrm{test}}$, with nine single-object specialist $\{\pi^j_{\mathrm{test}}\}_{j=1}^{9}$, each trained on the corresponding object's trajectories. 
Each single-object specialist is trained for 3k iterations. The unified policy is trained for $N_{\mathrm{obj}}\times 3$k iterations, where $N_{\mathrm{obj}}=9$, giving 27k iterations in total. Thus, the unified policy and the collection of nine specialists receive the same aggregate number of training iterations.
This comparison measures joint skill acquisition on a shared reference collection. All policies use the same observations, actions, reward, network architecture, and PPO configuration.
We do not tune hyperparameters for either the specialists or the unified policy.

\textbf{Results.}
As shown in Figure~\ref{fig:multi_object_comparison}, the unified policy achieves a pooled success rate of $95.3\%$, compared with $91.5\%$ for the specialists. 
Joint training matches specialist success on three objects and improves it on the other six, so the benefit is distributed across multiple objects rather than driven by a single easy category, and no object regresses. 
The pooled rates are weighted by the number of evaluation trajectories and therefore give greater weight to objects with more clips. 

\emph{Success and imitation fidelity dissociate.}
The specialists often attain slightly lower tracking errors while completing fewer interactions.
A specialist fits a single object's reference closely, whereas the unified policy allocates capacity across nine contact patterns and converges to a more conservative solution that trades pose fidelity for recovery margin.
Lower imitation error therefore does not imply more reliable interaction completion, which argues against using tracking error alone as the primary metric for contact-rich interaction.

\emph{Why joint training helps.}
We offer two mechanisms consistent with the evidence.
First, the approach and transport phases share whole-body structure across objects, such as walking to a pose, balancing the torso, and bearing a load, so clips from different objects act as mutual augmentation.
Second, contact patterns are shared across objects of similar shape: a small table and a large table are grasped in much the same way, as are a clothes stand and a floor lamp.
A specialist sees one geometry and can memorize a single grasp that fits it, whereas joint training exposes the policy to families of related shapes and pushes it toward selecting a grasp according to the object's geometry, which benefits to unseen interaction generalization.

\subsection{Architecture and Optimizer Ablation}
\label{sec:exp_ablation} 

\begin{figure}[h]
\centering
\includegraphics[width=\columnwidth]
{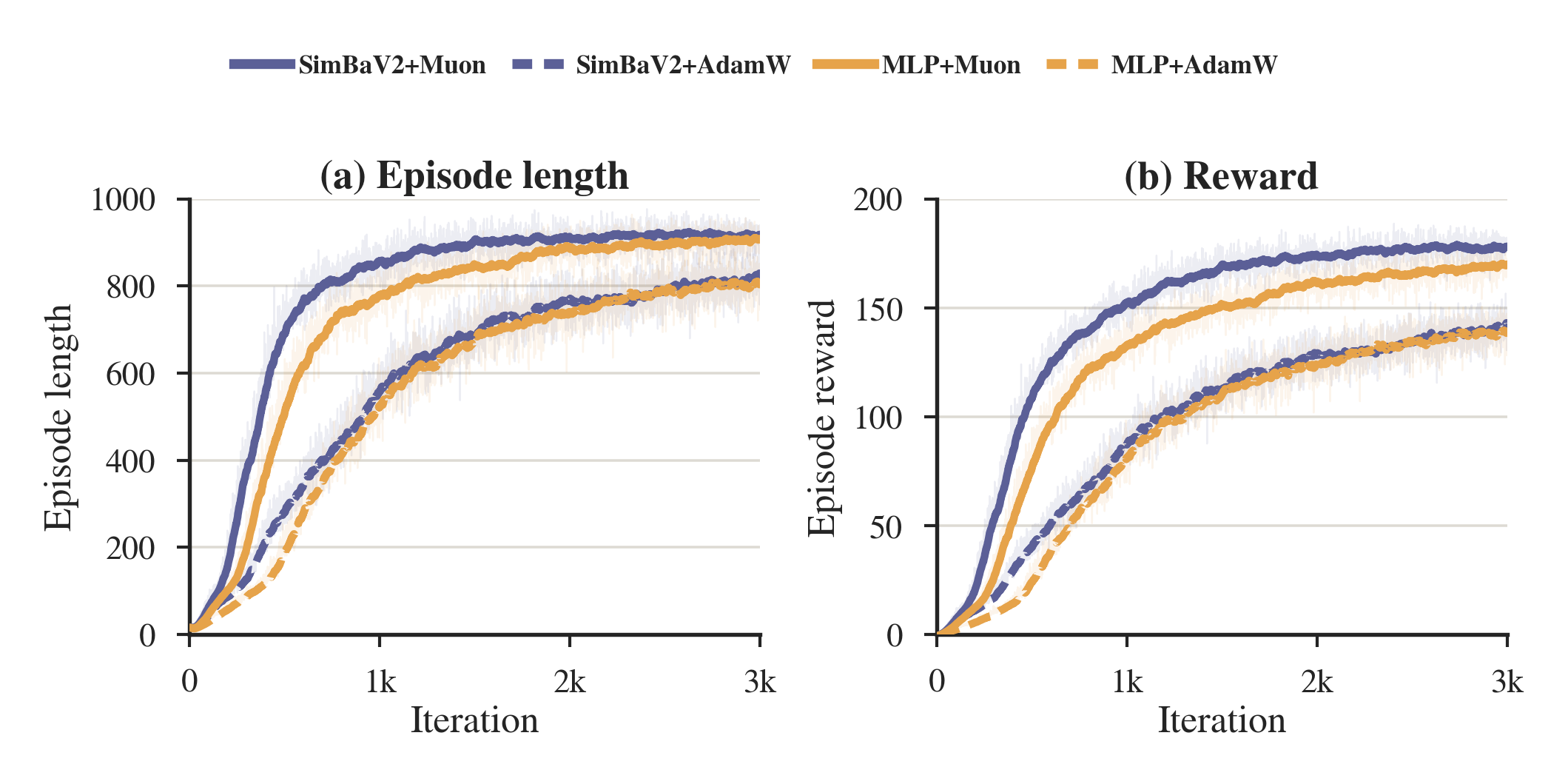}
\caption{\textbf{Sample-efficiency ablation on the small-table task.}
We compare four combinations under a common training budget of 3,000 iterations. Light curves show the raw mean episode length and reward, while dark curves show exponential moving averages with a span of 50 iterations.}
\label{fig:sample_efficiency}
\end{figure}

\textbf{Protocol.}
We evaluate all four combinations of MLP and SimBaV2 with AdamW and Muon on the small-table task under a common budget of 3,000 training iterations. Figure~\ref{fig:sample_efficiency} reports mean episode length and reward, with raw curves and exponential moving averages.

\textbf{Results.}
Both Muon configurations improve earlier than their AdamW counterparts, suggesting that the optimizer contributes the larger gain in sample efficiency in this comparison.
A plausible explanation is that Muon's orthogonalized momentum updates, so no single direction dominates a gradient update step.
This matters because the reward couples body tracking, object tracking, and finger contact, whose gradients differ substantially in scale: an update dominated by the body tracking term yields a policy that follows the reference pose without closing fingers, a local optimum that early training falls into easily.
SimBaV2 provides a smaller additional improvement.
Its hyperspherical normalization constrains weight and feature norms, which limits how far the policy moves in a single updates as the state distribution shifts at the onset of the contact, where the reward landscape changes most abruptly.
The two effects are complementary: the optimizer governs how the gradient is distributed across the coupled objectives, while the architecture governs how much the policy is allowed to change when the state distribution moves.

\section{Conclusions and Limitations}
We present \modelname{}, a framework that converts captured human-object interactions into whole-body dexterous loco-manipulation skills.
A reference-construction pipeline preserves task-relevant hand-object contacts across the morphology gap and completes the missing approach phases, and a unified contact- and geometry-aware policy, trained jointly over diverse object and multiple interaction clips, coordinates locomotion, whole-body balance, and finger-level grasping.
The unified policy completes $92.5\%$ success rate on trained interactions, and without any additional training, $65.0\%$ on sequences beyond training data distribution, overall produces $\sim$23h of physically executed rollouts.

However, several limitations remain.
Evaluation is carried out entirely in simulation: the policy consumes ground-truth odometry and object pose, so real-world deployment requires either onboard state estimation or distillation into a perception-based student, and the sim2real gap for contact-rich interaction is not accessed here.
Concern for embodiment, the Inspire hand is underactuated and the policy commands proximal joints only, which bounds the achievable finger-level fidelity, and heavy or highly articulated objects are outside the range covered by our references and domain randomizations.
Finally, the behaviors are reference-conditioned: the policy executes an interaction that must be supplied to it, and pairing it with a higher-level planner that selects and composes interaction, for instance, the humanoid VLA systems discussed in \ref{sec:h-vla}, remains future work.

\bibliographystyle{IEEEtran}
\bibliography{reference}

\begin{thebibliography}{10}
\providecommand{\url}[1]{#1}
\csname url@samestyle\endcsname
\providecommand{\newblock}{\relax}
\providecommand{\bibinfo}[2]{#2}
\providecommand{\BIBentrySTDinterwordspacing}{\spaceskip=0pt\relax}
\providecommand{\BIBentryALTinterwordstretchfactor}{4}
\providecommand{\BIBentryALTinterwordspacing}{\spaceskip=\fontdimen2\font plus
\BIBentryALTinterwordstretchfactor\fontdimen3\font minus
  \fontdimen4\font\relax}
\providecommand{\BIBforeignlanguage}[2]{{%
\expandafter\ifx\csname l@#1\endcsname\relax
\typeout{** WARNING: IEEEtran.bst: No hyphenation pattern has been}%
\typeout{** loaded for the language `#1'. Using the pattern for}%
\typeout{** the default language instead.}%
\else
\language=\csname l@#1\endcsname
\fi
#2}}
\providecommand{\BIBdecl}{\relax}
\BIBdecl

\bibitem{liu2025opt2skill}
F.~Liu, Z.~Gu, Y.~Cai, Z.~Zhou, H.~Jung, J.~Jang, S.~Zhao, S.~Ha, Y.~Chen,
  D.~Xu \emph{et~al.}, ``Opt2skill: Imitating dynamically-feasible whole-body
  trajectories for versatile humanoid loco-manipulation,'' \emph{IEEE Robotics
  and Automation Letters}, 2025.

\bibitem{weng2025hdmi}
H.~Weng, Y.~Li, N.~Sobanbabu, Z.~Wang, Z.~Luo, T.~He, D.~Ramanan, and G.~Shi,
  ``Hdmi: Learning interactive humanoid whole-body control from human videos,''
  \emph{arXiv preprint arXiv:2509.16757}, 2025.

\bibitem{yang2025omniretarget}
L.~Yang, X.~Huang, Z.~Wu, A.~Kanazawa, P.~Abbeel, C.~Sferrazza, C.~K. Liu,
  R.~Duan, and G.~Shi, ``Omniretarget: Interaction-preserving data generation
  for humanoid whole-body loco-manipulation and scene interaction,''
  \emph{arXiv preprint arXiv:2509.26633}, 2025.

\bibitem{zhao2025resmimic}
S.~Zhao, Y.~Ze, Y.~Wang, C.~K. Liu, P.~Abbeel, G.~Shi, and R.~Duan, ``Resmimic:
  From general motion tracking to humanoid whole-body loco-manipulation via
  residual learning,'' \emph{arXiv preprint arXiv:2510.05070}, 2025.

\bibitem{chen2026scenebot}
S.~Chen, S.~Zhao, Z.~Wu, J.~Li, G.~Shi, and C.~K. Liu, ``Scenebot:
  Contact-prompted general humanoid whole body tracking with
  scene-interaction,'' \emph{arXiv preprint arXiv:2606.27581}, 2026.

\bibitem{luo2024omnigrasp}
Z.~Luo, J.~Cao, S.~Christen, A.~Winkler, K.~Kitani, and W.~Xu, ``Omnigrasp:
  Grasping diverse objects with simulated humanoids,'' \emph{Advances in Neural
  Information Processing Systems}, vol.~37, pp. 2161--2184, 2024.

\bibitem{luo2025emergent}
Z.~Luo, C.~Tessler, T.~Lin, Y.~Yuan, T.~He, W.~Xiao, Y.~Guo, G.~Chechik,
  K.~Kitani, L.~Fan \emph{et~al.}, ``Emergent active perception and dexterity
  of simulated humanoids from visual reinforcement learning,'' \emph{arXiv
  preprint arXiv:2505.12278}, 2025.

\bibitem{tessler2025maskedmanipulator}
C.~Tessler, Y.~Jiang, E.~Coumans, Z.~Luo, X.~B. Peng, and G.~Chechik,
  ``Maskedmanipulator: Versatile whole-body control for loco-manipulation,'' in
  \emph{Proceedings of the SIGGRAPH Asia 2025 Conference Papers}, 2025, pp.
  1--11.

\bibitem{xu2025intermimic}
S.~Xu, H.~Y. Ling, Y.-X. Wang, and L.-Y. Gui, ``Intermimic: Towards universal
  whole-body control for physics-based human-object interactions,'' in
  \emph{Proceedings of the Computer Vision and Pattern Recognition Conference},
  2025, pp. 12\,266--12\,277.

\bibitem{xu2026interprior}
S.~Xu, S.~Schulter, M.~Ziyadi, X.~He, X.~Fei, Y.-X. Wang, and L.~Gui,
  ``Interprior: Scaling generative control for physics-based human-object
  interactions,'' \emph{arXiv preprint arXiv:2602.06035}, 2026.

\bibitem{wu2025human}
Z.~Wu, J.~Li, P.~Xu, and C.~K. Liu, ``Human-object interaction from human-level
  instructions,'' in \emph{2025 IEEE/CVF International Conference on Computer
  Vision (ICCV)}.\hskip 1em plus 0.5em minus 0.4em\relax IEEE, 2025, pp.
  11\,176--11\,186.

\bibitem{rempe2026kimodo}
D.~Rempe, M.~Petrovich, Y.~Yuan, H.~Zhang, X.~B. Peng, Y.~Jiang, T.~Wang,
  U.~Iqbal, D.~Minor, M.~de~Ruyter \emph{et~al.}, ``Kimodo: Scaling
  controllable human motion generation,'' \emph{arXiv preprint
  arXiv:2603.15546}, 2026.

\bibitem{luo2023perpetual}
Z.~Luo, J.~Cao, K.~Kitani, W.~Xu \emph{et~al.}, ``Perpetual humanoid control
  for real-time simulated avatars,'' in \emph{Proceedings of the IEEE/CVF
  International Conference on Computer Vision}, 2023, pp. 10\,895--10\,904.

\bibitem{Zakka_Mink_Python_inverse_2026}
\BIBentryALTinterwordspacing
K.~Zakka, ``{Mink: Python inverse kinematics based on MuJoCo},'' Feb. 2026.
  [Online]. Available: \url{https://github.com/kevinzakka/mink}
\BIBentrySTDinterwordspacing

\bibitem{he2024omnih2o}
T.~He, Z.~Luo, X.~He, W.~Xiao, C.~Zhang, W.~Zhang, K.~Kitani, C.~Liu, and
  G.~Shi, ``Omnih2o: Universal and dexterous human-to-humanoid whole-body
  teleoperation and learning,'' \emph{arXiv preprint arXiv:2406.08858}, 2024.

\bibitem{wu2026toporetarget}
J.~Wu, S.~Yao, G.~He, X.~Liu, Z.~Zeng, X.~Jiang, H.~Yang, W.~Zhang, and
  H.~Zhao, ``Toporetarget: Interaction-preserving retargeting for dexterous
  manipulation,'' \emph{arXiv preprint arXiv:2606.16272}, 2026.

\bibitem{feng2026minimalist}
Y.~Feng, N.~Leung, J.~Wang, L.~Yang, H.~Qi, and P.~Culbertson, ``A minimalist
  retargeting-guided reinforcement learning recipe for dexterous
  manipulation,'' \emph{arXiv preprint arXiv:2607.11874}, 2026.

\bibitem{wu2026reforce}
Y.~Wu, L.~Zeng, C.~Jing, J.~Ye, and X.~Wang, ``Reforce: Learning force-aware
  retargeting for dexterous manipulation,'' \emph{arXiv preprint
  arXiv:2608.15560}, 2026.

\bibitem{shao2026synmandex}
Y.~Shao, Z.~Chen, W.~Lin, M.~Zhou, T.~Chen, X.~Yang, Y.~Chi, and Y.~Mu,
  ``Synmandex: Synthesizing human-like dexterous grasps from synthetic human
  pre-grasps,'' \emph{arXiv preprint arXiv:2606.09798}, 2026.

\bibitem{zhu2026learning}
X.~Zhu, Z.~Liu, S.~Jain, C.~Li, M.~Noori, M.~A. Lin, H.~Zhao, J.~Welsh,
  M.~Verghese, W.~Liu \emph{et~al.}, ``Learning dexterous manipulation using
  contact wrench guidance from human demonstration,'' \emph{arXiv preprint
  arXiv:2607.00033}, 2026.

\bibitem{chen1dex}
R.~Chen, F.~Ruan, L.~Cao, Z.~Wang, B.~Xu, S.~Tong, J.~Liu, M.~Pei, C.~Zhang,
  W.~Xing \emph{et~al.}, ``Dex-x: Learning visual-tactile dexterous
  manipulation from human videos with simulated interaction,''
  \emph{challenge}, vol.~1, no.~29, p.~30.

\bibitem{wang2026motionbricks}
T.~Wang, O.~Dionne, M.~De~Ruyter, D.~Minor, D.~Rempe, K.~Zhao, M.~Petrovich,
  Y.~Yuan, C.~Li, Z.~Luo \emph{et~al.}, ``Motionbricks: Scalable real-time
  motions with modular latent generative model and smart primitives,''
  \emph{ACM Transactions on Graphics (TOG)}, vol.~45, no.~4, pp. 1--22, 2026.

\bibitem{zhao2026ardy}
K.~Zhao, M.~Petrovich, H.~Zhang, T.~Wang, S.~Tang, and D.~Rempe, ``Ardy:
  Autoregressive diffusion with hybrid representation for interactive human
  motion generation,'' \emph{arXiv preprint arXiv:2607.08741}, 2026.

\bibitem{chen2025hand}
S.~Chen, Y.~Ye, Z.-a. Cao, J.~Lew, P.~Xu, and C.~K. Liu, ``Hand-eye autonomous
  delivery: Learning humanoid navigation, locomotion and reaching,''
  \emph{arXiv preprint arXiv:2508.03068}, 2025.

\bibitem{arnaud2025locate}
S.~Arnaud, P.~McVay, A.~Martin, A.~Majumdar, K.~M. Jatavallabhula, P.~Thomas,
  R.~Partsey, D.~Dugas, A.~Gejji, A.~Sax \emph{et~al.}, ``Locate 3d: Real-world
  object localization via self-supervised learning in 3d,'' \emph{arXiv
  preprint arXiv:2504.14151}, 2025.

\bibitem{liu2025compass}
W.~Liu, H.~Zhao, C.~Li, Y.~Deng, J.~Biswas, S.~Pouya, and Y.~Chang, ``Compass:
  Cross-embodiment mobility policy via residual rl and skill synthesis,''
  \emph{arXiv preprint arXiv:2502.16372}, 2025.

\bibitem{he2025viral}
T.~He, Z.~Wang, H.~Xue, Q.~Ben, Z.~Luo, W.~Xiao, Y.~Yuan, X.~Da,
  F.~Casta{\~n}eda, S.~Sastry \emph{et~al.}, ``Viral: Visual sim-to-real at
  scale for humanoid loco-manipulation,'' \emph{arXiv preprint
  arXiv:2511.15200}, 2025.

\bibitem{Taouil2026MotionDiscoMD}
I.~Taouil, M.~Ciebelski, S.~Omar, H.~Zhao, A.~Dai, A.~M. Johnson, and
  M.~Khadiv, ``Motiondisco: Motion discovery for extreme humanoid
  loco-manipulation,'' 2026.

\bibitem{li2026vaic}
D.~Li, Q.~Wu, X.~Chen, L.~Li, Y.~Lin, S.~Wu, G.~Zhang, M.~Zhou, D.~Xiang,
  Q.~Zhang \emph{et~al.}, ``Vaic: Vision-guided humanoid agile object
  interaction control via decoupled commands,'' \emph{arXiv preprint
  arXiv:2606.09286}, 2026.

\bibitem{wang2025physhsi}
H.~Wang, W.~Zhang, R.~Yu, T.~Huang, J.~Ren, F.~Jia, Z.~Wang, X.~Niu, X.~Chen,
  J.~Chen \emph{et~al.}, ``Physhsi: Towards a real-world generalizable and
  natural humanoid-scene interaction system,'' \emph{arXiv preprint
  arXiv:2510.11072}, 2025.

\bibitem{he2026ultra}
X.~He, S.~Xu, X.~Li, R.~Dong, L.~Bian, Y.-X. Wang, and L.-Y. Gui, ``Ultra:
  Unified multimodal control for autonomous humanoid whole-body
  loco-manipulation,'' \emph{arXiv preprint arXiv:2603.03279}, 2026.

\bibitem{byrd2026adaptmanip}
M.~Byrd, D.~Baek, K.~Garg, H.~Jung, D.~Cho, M.~Sorokin, R.~Wright, and S.~Ha,
  ``Adaptmanip: Learning adaptive whole-body object lifting and delivery with
  online recurrent state estimation,'' \emph{arXiv preprint arXiv:2602.14363},
  2026.

\bibitem{yin2025visualmimic}
S.~Yin, Y.~Ze, H.-X. Yu, C.~K. Liu, and J.~Wu, ``Visualmimic: Visual humanoid
  loco-manipulation via motion tracking and generation,'' \emph{arXiv preprint
  arXiv:2509.20322}, 2025.

\bibitem{xie2026grailgeneratinghumanoidlocomanipulation}
\BIBentryALTinterwordspacing
T.~Xie, H.~Zhang, J.~Park, Z.~Wang, B.~Wen, J.~Li, X.~Li, Q.~Ben, H.~Weng,
  Y.~Ye, D.~Minor, T.~Wang, C.~Jiang, S.~Fidler, J.~Kautz, L.~Fan, Y.~Zhu,
  Z.~Luo, U.~Iqbal, and Y.~Yuan, ``Grail: Generating humanoid loco-manipulation
  from 3d assets and video priors,'' 2026. [Online]. Available:
  \url{https://arxiv.org/abs/2606.05160}
\BIBentrySTDinterwordspacing

\bibitem{team2026kling}
K.~Team, J.~Chen, Y.~Ding, Z.~Fang, K.~Gai, K.~He, X.~He, J.~Hua, M.~Lao, X.~Li
  \emph{et~al.}, ``Kling-motioncontrol technical report,'' \emph{arXiv preprint
  arXiv:2603.03160}, 2026.

\bibitem{lin2026humanoidmimicgen}
K.~Lin, A.~Mandlekar, C.~R. Garrett, N.~Chernyadev, Y.~Fang, R.~Ding, Y.~Xie,
  J.~Tran, L.~Fan, and Y.~Zhu, ``Humanoidmimicgen: Data generation for
  loco-manipulation via whole-body planning,'' \emph{arXiv preprint
  arXiv:2605.27724}, 2026.

\bibitem{wang2026vlk}
Y.-J. Wang, J.~Li, S.~Chen, T.~E. Truong, P.~Xu, P.~Abbeel, R.~Duan,
  K.~Sreenath, A.~Kanazawa, C.~Sferrazza \emph{et~al.}, ``Vlk: Learning
  humanoid loco-manipulation from synthetic interactions in reconstructed
  scenes,'' \emph{arXiv preprint arXiv:2606.30645}, 2026.

\bibitem{kim2024openvla}
M.~J. Kim, K.~Pertsch, S.~Karamcheti, T.~Xiao, A.~Balakrishna, S.~Nair,
  R.~Rafailov, E.~Foster, G.~Lam, P.~Sanketi \emph{et~al.}, ``Openvla: An
  open-source vision-language-action model,'' \emph{arXiv preprint
  arXiv:2406.09246}, 2024.

\bibitem{black2024pi_0}
K.~Black, N.~Brown, D.~Driess, A.~Esmail, M.~Equi, C.~Finn, N.~Fusai, L.~Groom,
  K.~Hausman, B.~Ichter \emph{et~al.}, ``{$\pi_0$}: A vision-language-action
  flow model for general robot control,'' \emph{arXiv preprint
  arXiv:2410.24164}, 2024.

\bibitem{intelligence2025pi_}
P.~Intelligence, K.~Black, N.~Brown, J.~Darpinian, K.~Dhabalia, D.~Driess,
  A.~Esmail, M.~Equi, C.~Finn, N.~Fusai \emph{et~al.}, ``{$\pi_{0.5}$}: a
  vision-language-action model with open-world generalization,'' \emph{arXiv
  preprint arXiv:2504.16054}, 2025.

\bibitem{bjorck2025gr00t}
J.~Bjorck, F.~Casta{\~n}eda, N.~Cherniadev, X.~Da, R.~Ding, L.~Fan, Y.~Fang,
  D.~Fox, F.~Hu, S.~Huang \emph{et~al.}, ``Gr00t n1: An open foundation model
  for generalist humanoid robots,'' \emph{arXiv preprint arXiv:2503.14734},
  2025.

\bibitem{jiang2025wholebodyvla}
H.~Jiang, J.~Chen, Q.~Bu, L.~Chen, M.~Shi, Y.~Zhang, D.~Li, C.~Suo, C.~Wang,
  Z.~Peng \emph{et~al.}, ``Wholebodyvla: Towards unified latent vla for
  whole-body loco-manipulation control,'' \emph{arXiv preprint
  arXiv:2512.11047}, 2025.

\bibitem{bai2026hex}
S.~Bai, M.~Li, X.~Lv, J.~Wang, X.~Wang, F.~Liao, C.~Hou, L.~Gu, W.~Zhou, K.~Wu
  \emph{et~al.}, ``Hex: Humanoid-aligned experts for cross-embodiment
  whole-body manipulation,'' \emph{arXiv preprint arXiv:2604.07993}, 2026.

\bibitem{wei2026psi_0}
S.~Wei, H.~Jing, B.~Li, Z.~Zhao, J.~Mao, Z.~Ni, S.~He, J.~Liu, X.~Liu, K.~Kang
  \emph{et~al.}, ``{{$\Psi_0$}}: An open foundation model towards universal
  humanoid loco-manipulation,'' \emph{arXiv preprint arXiv:2603.12263}, 2026.

\bibitem{li2026omega}
Z.~Li, Z.~Zhang, Y.~Wei, W.~Zhang, X.~Yuan, P.~Zhi, G.~Li, X.~Guo, F.~Gao,
  J.~Yang \emph{et~al.}, ``$omega $-0: A latent predictive world action model
  for concurrent humanoid loco-manipulation,'' \emph{arXiv preprint
  arXiv:2608.06375}, 2026.

\bibitem{li2023objectmotionguidedhuman}
\BIBentryALTinterwordspacing
J.~Li, J.~Wu, and C.~K. Liu, ``Object motion guided human motion synthesis,''
  2023. [Online]. Available: \url{https://arxiv.org/abs/2309.16237}
\BIBentrySTDinterwordspacing

\bibitem{araujo2025retargeting}
J.~P. Araujo, Y.~Ze, P.~Xu, J.~Wu, and C.~K. Liu, ``Retargeting matters:
  General motion retargeting for humanoid motion tracking,'' \emph{arXiv
  preprint arXiv:2510.02252}, 2025.

\bibitem{ferrari1992planning}
C.~Ferrari, J.~Canny \emph{et~al.}, ``Planning optimal grasps,'' in
  \emph{Proceedings., 1992 IEEE International Conference on Robotics and
  Automation, 1992.}, vol.~3.\hskip 1em plus 0.5em minus 0.4em\relax IEEE,
  1992, pp. 2290--2295.

\bibitem{schulman2017proximal}
J.~Schulman, F.~Wolski, P.~Dhariwal, A.~Radford, and O.~Klimov, ``Proximal
  policy optimization algorithms,'' \emph{arXiv preprint arXiv:1707.06347},
  2017.

\bibitem{peng2018deepmimic}
X.~B. Peng, P.~Abbeel, S.~Levine, and M.~Van~de Panne, ``Deepmimic:
  Example-guided deep reinforcement learning of physics-based character
  skills,'' \emph{ACM Transactions On Graphics (TOG)}, vol.~37, no.~4, pp.
  1--14, 2018.

\bibitem{liao2025beyondmimic}
Q.~Liao, T.~E. Truong, X.~Huang, Y.~Gao, G.~Tevet, K.~Sreenath, and C.~K. Liu,
  ``Beyondmimic: From motion tracking to versatile humanoid control via guided
  diffusion,'' \emph{arXiv preprint arXiv:2508.08241}, 2025.

\bibitem{lee2025simba}
H.~Lee, D.~Hwang, D.~Kim, H.~Kim, J.~J. Tai, K.~Subramanian, P.~Wurman,
  J.~Choo, P.~Stone, and T.~Seno, ``Simba: Simplicity bias for scaling up
  parameters in deep reinforcement learning,'' in \emph{International
  Conference on Learning Representations}, vol. 2025, 2025, pp.
  46\,816--46\,848.

\bibitem{liu2025muon}
J.~Liu, J.~Su, X.~Yao, Z.~Jiang, G.~Lai, Y.~Du, Y.~Qin, W.~Xu, E.~Lu, J.~Yan
  \emph{et~al.}, ``Muon is scalable for llm training,'' \emph{arXiv preprint
  arXiv:2502.16982}, 2025.

\end{thebibliography}

\end{document}